\documentclass[10pt,twocolumn,letterpaper]{article}

\usepackage{cvpr}
\usepackage{times}
\usepackage{epsfig}
\usepackage{graphicx}
\usepackage{amsmath}
\usepackage{amssymb}
\usepackage[noend]{algpseudocode}

\usepackage{algorithmicx,algorithm}
\newcommand{\tabincell}[2]{\begin{tabular}{@{}#1@{}}#2\end{tabular}}
\usepackage{multirow}
\usepackage{makecell}
\usepackage{microtype}
\usepackage{subfigure}
\usepackage{booktabs} 

\usepackage[pagebackref=true,breaklinks=true,letterpaper=true,colorlinks,bookmarks=false]{hyperref}

\cvprfinalcopy 

\def\cvprPaperID{****} 

\ifcvprfinal\fi
\begin{document}

\title{A Transductive Multi-Head Model for Cross-Domain Few-Shot Learning}

\author{
	Jianan Jiang,\; Zhenpeng Li, \; Yuhong Guo, \; Jieping Ye\\	
AI Tech, DiDiChuXing  
}

\maketitle

\begin{abstract}
In this paper, we present a new method, Transductive Multi-Head Few-Shot learning (TMHFS),
to address the Cross-Domain Few-Shot Learning (CD-FSL) challenge. 
The TMHFS method extends the Meta-Confidence Transduction (MCT) and Dense Feature-Matching Networks (DFMN)
method~\cite{kye2020transductive} 
by introducing a new prediction head, i.e, an instance-wise global classification network 
based on semantic information, 
after the common feature embedding network. 
We train the embedding network with the multiple heads, i.e,, the MCT loss, the DFMN loss and 
the semantic classifier loss, simultaneously in the source domain. 
For the few-shot learning in the target domain, 
we first perform fine-tuning on the embedding network with only the semantic global classifier
and the support instances,
and then use the MCT part to predict labels of the query set with the fine-tuned embedding network.
Moreover, we further exploit data augmentation techniques during the fine-tuning and test stages 
to improve the prediction performance. 
The experimental results demonstrate that the proposed methods greatly outperform the strong baseline, fine-tuning,
on four different target domains. 
\end{abstract}

\section{Introduction}

The task of few-shot classification only has a few labeled instances from each class for training. 
To address this insufficiency of labeled data, 
cross domain few-shot learning aims to exploit the abundant labeled data in a different source domain. 
The main challenge of cross-domain few-shot learning 
lies in the cross domain divergences in both the input data space and the output label space;
that is, not only the classes of the two domains are entirely different, 
their input images have very different appearances as well. 
It has been shown in~\cite{Guo2019A} that
in such a cross-domain learning scenario
a simple fine-tuning method can outperform many classic few-shot learning techniques.

In this paper, we present a new method, Transductive Multi-Head Few-Shot learning (TMHFS),
to address the cross-domain few-shot learning challenge. 
TMHFS is based on the Meta-Confidence Transduction (MCT) and Dense Feature-Matching Networks (DFMN)
method developed in~\cite{kye2020transductive}.
It extends the transductive model in ~\cite{kye2020transductive} 
by adding an instance-wise global classification network based on the semantic information, 
after the common feature embedding network as a new prediction ``head''.
The method consists of three stages: training stage, fine-tuning stage and testing stage. 
During the training stage, 
we train the embedding network with the multiple prediction heads, i.e,,
the distance based instance meta-train classifier, the pixel wise classifier, and the semantic information
based global-wise classifier, in the source domain. 
In the fine-tuning stage, we fine-tune the model 
using only the semantic global-wise classifier and the support instances in the target domain.
Finally, in the testing stage 
we use the MCT part, i.e., the meta-trained instance classifier, to predict labels of the query set with the fine-tuned embedding network.
Moreover, we further incorporate data augmentation techniques during the fine-tuning and test stages 
to improve the prediction performance. 
The experimental results demonstrate that the proposed methods greatly outperform the strong baseline, fine-tuning,
on the four target domains. 
We conduct experiments on four target domains (CropDisease, EuroSAT, ISIC, ChestX) using miniImageNet as the source domain. The results show our methods consistently outperform the strong fine-tuning baseline method
and achieve notable performance gains.

\section{Our Approach }

\begin{figure*}[htbp]
\centering
\includegraphics[width=0.85\linewidth]{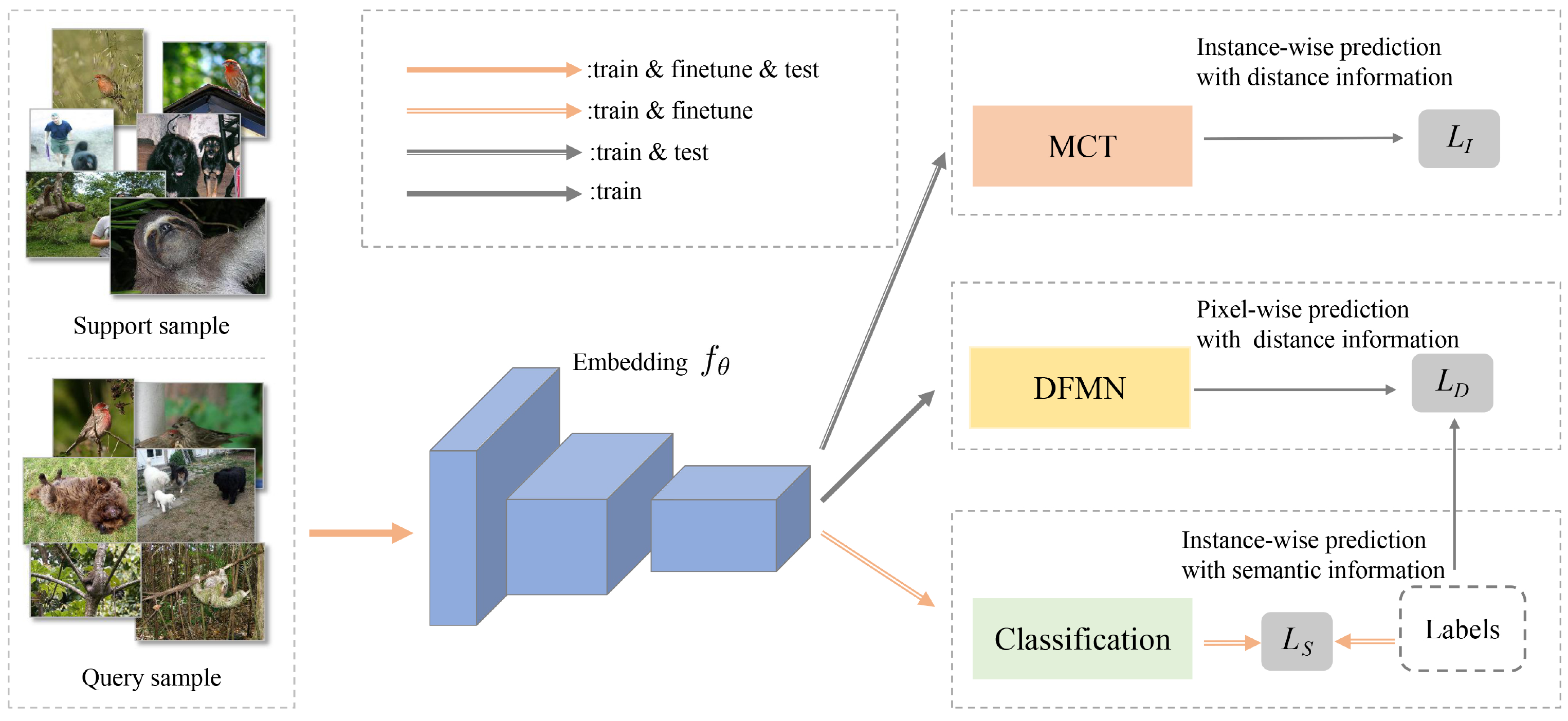}
\caption{
	The proposed transductive multi-head few-shot (TMHFS) learning model. }
\label{fig:fig1}
\end{figure*}

\subsection{Problem Statement}

In cross-domain few-shot learning setting, we have a source domain 
$S=\left \{X_{s},Y_{s}\right\}$ from a total $\mathcal{C}_g$ classes
and a target domain $T=\left\{X_{t},Y_{t} \right\}$ from a set of totally different classes. 
The two domains have different marginal distributions in the input feature space,
and disjoint output class sets. 
The source domain has abundant labeled instances for training,
while a set of C-way N-shot few-shot learning tasks are 
sampled for meta-training ($C < K$).
In each few-shot task, 
$C$ novel classes are randomly selected from the entire target class set.
From each class, $N$ and $M$ instances are randomly selected as the support set and query set; 
i.e., ${S} = \left \{  \left ( \textbf{x}_1,y_1 \right ), \left ( \textbf{x}_2,y_2 \right ),...,\left ( \textbf{x}_{CN},y_{CN} \right ) \right \}$ and ${Q} = \left \{  \left ( \tilde{\textbf{x}}_1,\tilde{y}_1 \right ), \left ( \tilde{\textbf{x}}_2,\tilde{y}_2 \right ),...,\left ( \tilde{\textbf{x}}_{CM},\tilde{y}_{CM} \right ) \right \}$. 
Similarly, such few-shot tasks from the target domain are used for few-shot learning evaluation. 

\subsection{Base Model: Transductive Few-Shot Learning with Meta-Learned Confidence}

Our proposed model is illustrated in Figure~\ref{fig:fig1}. 
It adopts the Transductive Few-Shot Learning with Meta-Learned Confidence method 
from \cite{kye2020transductive} as our base model, for which we provide an overview in this subsection. 

The transductive base model has two prediction heads, the instance-wise Meta-Confidence Transduction (MCT)
and the pixel-wise Dense Feature-Matching Network (DFMN), which share the same feature extraction network $f_\theta$.
The MCT uses distance based prototype classifier to make prediction for the query instances, 
\begin{align}
p\left ( \tilde{y}_l =c|\tilde{\textbf{x}},{S};\theta ,\phi \right )=\frac{\exp \left ( -d_\phi \left ( f_\theta \left (\tilde{\textbf{x}}  \right ), P_{c}^{T} \right ) \right )}{\sum_{{c}'=1}^{C}\exp \left (-d_\phi \left ( f_\theta \left (\tilde{\textbf{x}}  \right ), P_{{c}'}^{T} \right ) \right )}
\label{eq:p_i}
\end{align}
where $P_c$ denotes the prototype vector for the $c$-th class in the given C-way few-shot training task, 
and $\tilde{y}_l\in \left \{ 1,2,...C \right \}$ denotes the C-way {\em local} label of the query instance. 
$d_\phi$ denotes an Euclidean distance function with normalization and input-dependent length-scaling with parameter $\phi$. 
The prototype vector of each class $P_{c}$ can be initialized on the support set 
and then iteratively updated using the query set:
\begin{align}
P_c^0 &= \frac{1}{{S}_c}\sum_{\textbf{x} \in {S}_c}f_\theta \left ( \textbf{x} \right )
	\label{eq:T1}
\\
q_c^{t-1}(\tilde{\textbf{x} })&=\frac{\exp \left ( -d_\phi \left ( f_\theta \left (\tilde{\textbf{x}}  \right ), P_{c}^{t-1} \right ) \right )}{\sum_{{c}'=1}^{C}\exp \left (-d_\phi \left ( f_\theta \left (\tilde{\textbf{x}}  \right ), P_{{c}'}^{t-1} \right ) \right )}
\end{align}
\begin{align}
P_c^t&=\frac{\sum_{\textbf{x} \in {S}_c}1\cdot f_\theta \left ( \textbf{x} \right )+\sum_{\tilde{\textbf{x}} \in Q_x}q_c^{t-1}\left ( \tilde{\textbf{x}} \right )\cdot f_\theta \left ( \tilde{\textbf{x}} \right )}{\sum_{\textbf{x} \in {S}_c}1+\sum_{\tilde{\textbf{x}} \in {Q}_x}q_c^{t-1}\left ( \tilde{\textbf{x}} \right )}
	\label{eq:T3}
\end{align}
where ${S}_c$ denotes the set of support samples belonging to each class $c\in\{ 1,...,C\}$,
and $t=1,...,T$ denotes the number of transduction iterations. 
$T=1$ is used when training the model in the source domain 
and $T=10$ is used when testing in the target domain.

The pixel-wise Dense Feature-Matching Network (DFMN) is used solely in the training stage. 
It extracts $K$-dimensional feature vector at the location i, $f^i_\theta({\bf x})$, 
and uses a set of global prototypes for each class, 
$\omega =\left \{ \textbf{w}_c \in \mathbb{R}^K | c=1,...,\mathcal{C}_g\right \}$.
For each pixel $i \in \left \{ \left ( 1,1 \right ),...,\left ( H,W \right ) \right \}$ the prediction is:
\begin{equation}
	p^i\left ( \tilde{y}_{g} |\tilde{\textbf{x}}; \theta ,\omega  \right )=\frac{\exp \left ( -d\left ( f^i_\theta\left ( \tilde{\textbf{x}} \right ),\textbf{w}_{\tilde{y}_{g} } \right ) \right )}{\sum_{c=1}^{\mathcal{C}_{g}}\exp \left ( -d\left ( f^i_\theta\left ( \tilde{\textbf{x}} \right ),\textbf{w}_{c} \right ) \right )}
\end{equation}

Training in the source domain combines both MCT and DFMN. 
Given a set of $C$-way $N$-shot tasks, $\tau$, sampled from a task distribution $p(\tau)$ in the source domain, 
the training is conducted using the following combination loss: 
\begin{equation}
L\!\left ( \theta, \phi, \omega \right )\!=\!\mathbb{E}_{p\left ( \tau  \right )}\!\left [ \lambda L_{I }^{\tau}\!\left ( \theta ,\phi  \right )\!+\!\frac{1}{H\times W}\sum_{i}^{H\times W}L_{D}^{\tau,i}\!\left ( \theta ,\omega  \right )\right ]
\label{eq:loss_t}
\end{equation}
where $L_{I}$ denotes the instance-wise loss of the Meta-Confidence Transduction (MCT) prediction, such that
\begin{equation}
L_{I }^{\tau}\left ( \theta ,\phi  \right )=\frac{1}{\left | Q^\tau \right |}\sum_{\left ( \tilde{\textbf{x}},\tilde{y }_l\right ) \in Q^\tau} -\log p (\tilde{y}_l \mid \tilde{\textbf{x}},S^\tau; \theta , \phi ) 
\label{eq:loss_i}
\end{equation}
$L_{D}$ denotes the pixel-wise loss of the Dense Feature-Matching Networks (DFMN) prediction:
\begin{equation}
L_{D}^{\tau,i}\left ( \theta ,\omega  \right )=\frac{1}{\left | {Q}^\tau \right |}\sum_{\left ( \tilde{\textbf{x}},\tilde{y}_{g}\right ) \in {Q}^\tau} -\log p (\tilde{y}_{g} \mid \tilde{\textbf{x}},{S}^\tau; \theta , \omega ) 
\label{eq:loss_d}
\end{equation}

\subsection{Transductive Multi-Head Few-Shot Learning}
We extend the transductive base model above by 
adding a new global instance-wise prediction head, $f_\delta$, 
based on the extracted semantic information with $f_\theta$.
For this prediction head, we consider the global classification problem over all the $\mathcal{C}_g$ classes. 
As shown in Figure~\ref{fig:fig1},
both support set and query set are used as training input for this branch, such that:
\begin{equation}
\begin{aligned}
p\left ( y_{g} | \textbf{x}; \theta ,\delta  \right ) &= f_\delta \left ( f_\theta \left ( \textbf{x} \right ) \right ),
	\forall ({\bf x}, y_g)\in S, \\
	p\left ( \tilde{y}_{g} |\tilde{\textbf{x}}; \theta ,\delta  \right ) &= f_\delta \left ( f_\theta \left ( \tilde{\textbf{x}} \right ) \right ),  \forall ({\bf \tilde{x}}, \tilde{y}_g)\in Q,
\end{aligned}
\end{equation}
where $f_\delta$ is a full connected one layer neural network with soft-max classification, 
$\delta$ denotes its parameters. 
The instance-wise loss from this semantic branch 
over each training task
can be written as follows:
\begin{equation}
\begin{aligned}
	L_{\mathcal{S}}^{\tau}\left ( \theta , \delta   \right )=\frac{-1}{\left | {Q}^\tau \right |
+\left | {S}^\tau \right |}  
	\left(\!\!
	\begin{array}{l}	
	\sum\limits_{( \tilde{\textbf{x}},\tilde{y }_{g}) \in {Q}^\tau}  \log p (\tilde{y}_{g} \mid \tilde{\textbf{x}}; \theta , \delta )+\\ 
\sum\limits_{( \textbf{x},y_{g}) \in {S^\tau}}  \log p (y_{g} \mid \textbf{x}; \theta , \delta )
	\end{array}\!\!\right)	
\end{aligned}
\label{eq:loss_s}
\end{equation}
Below we present the three stages 
(training stage, fine-tuning stage, and testing stage) 
 of the proposed TMHFS method for cross-domain few-shot learning. 

\paragraph{Training stage.} 
The purpose of training is to pre-train an embedding model 
$f_{\theta}$ (i.e., the feature extractor) in the source domain. 
For the TMHFS model, we perform training on the sampled few-shot tasks
by combing the losses in Eq.(\ref{eq:loss_i}), Eq.(\ref{eq:loss_d}), 
and Eq.(\ref{eq:loss_s}) from the three prediction heads: 
\begin{equation}
\begin{aligned}
L\left ( \theta, \phi, \omega, \delta  \right )=\mathbb{E}_{p\left ( \tau  \right )}
\left[ 
\begin{array}{l}
	\lambda L_{I }^{\tau}\left ( \theta ,\phi  \right )+\alpha  L_{S}^{\tau}\left ( \theta , \delta   \right )\\[1ex]
+\frac{1}{H\times W}\sum_{i}^{H\times W}L_{D}^{\tau,i}\left ( \theta ,\omega  \right )
\end{array}
	\right]
\end{aligned}
\label{eq:loss_all}
\end{equation}
where $\lambda$ and $\alpha$ are trade-off parameters.

\paragraph{Fine-tuning stage.} 
Given a few-shot learning task in the target domain, 
we fine-tune the embedding model 
$f_{\theta}$ on the support set 
by using only the instance-wise prediction head $f_\delta$,
aiming to adapt $f_\theta$ to the target domain data.
Specifically, starting with the pre-trained model parameters,
we minimize the following loss function
in terms of $\theta$ and $\delta$ on the labeled support instances
in the target domain: 
\begin{equation}
L_{S}\left ( \theta , \delta \right )=\frac{1}{\left| {S} \right| }\sum_{\left ( \textbf{x},y \right ) \in {S}} - \log p \left(y \mid \textbf{x}; \theta , \delta \right)
\end{equation}

\paragraph{Testing stage.}
After fine-tuning, we use only the instance-wise MCT prediction head 
to predict the labels of the query instances in the target domain.
The prediction is conducted using
Eq.(\ref{eq:p_i}) after performing the transductive steps (Eq.(\ref{eq:T1})--Eq.(\ref{eq:T3})) for T = 10 iterations.

\subsection{Data Augmentation}
To mitigate the insufficiency of the labeled data in the target domain 
and increase the robustness of prediction,
we further incorporate some data augmentation operations such as image scaling, resized crop, horizontal flip,
rotation, and image jitter, to augment the support and query instances in the target domain.
The details of different augmentation operations are shown in Table \ref{table:Augmentation}. 
Specifically, we apply $n_A$ different data augmentation methods or their combinations to each image
in the support set and query set and generate augmented sets: 
${S}^A = \left \{{S}_1,{S}_2,..,{S}_{n_A}  \right \}$ and ${Q}^A = \left \{{Q}_1,{Q}_2,..,{Q}_{n_A}  \right \}$, 
such that 
\begin{equation}
\begin{aligned}
{S}_i &\!=\! \left \{\!  \left ( A_i\!\left( \textbf{x}_1\right),y_1 \right )\!, \left ( A_i\!\left(  \textbf{x}_2\right ), y_2 \right )\!,...,\left (  A_i\!\left( \textbf{x}_{CN}\right ), y_{CN,} \right ) \!\right \}
\\
{Q}_i&\!= \!\left \{\!  \left ( A_i\!\left( \tilde{\textbf{x}}_1\right ),\tilde{y}_1 \right )\!, \left ( A_i\!\left( \tilde{\textbf{x}}_2\right ),\tilde{y}_2 \right )\!,...,\left ( A_i\!\left( \tilde{\textbf{x}}_{CM}\right )\!,\tilde{y}_{CM,} \right ) \!\right \}
\end{aligned}
\end{equation}
where $A_i$ denotes an augmentation function. 
Then 
at the fine-tuning stage, we fine-tune $f_\theta$ on the augmented support set $S^A$
by minimizing the following loss:
\begin{equation}
\begin{aligned}
L_{S_{}}\!\left ( \theta , \delta \right )\!=\!\frac{1}{n_A }\!\sum_{i=1}^{n_A}
\frac{1}{\left|\! {S}_i \!\right| }\sum_{\left ( \!A_i\left(\textbf{x} \right ),y \!\right ) \in {S}_i}
\!- \!\log p \!\left(y \!\mid\! A_i\!\left(\textbf{x} \right )\!; \theta , \!\delta \right)
\end{aligned}
\end{equation}
At the testing stage, we use the MCT head to perform 
prediction on each pair of sets, $(S_i, Q_i)$, separately from other sets.
Then the final prediction result over each image can be determined 
as the average of the multiple predictions obtained from its augmented variants:
\begin{equation}
p\left ( \tilde{y} =c|\tilde{\textbf{x}},{S}^A;\theta ,\phi \right )=\frac{1}{n_A}\sum_{i=1}^{n_A}p\left ( \tilde{y} =c|A_i\left( \tilde{ \textbf{x}}\right),{S}_i;\theta ,\phi \right )
\end{equation}

\section{Experiments }

\begin{table*}[h]
\begin{center}
\caption{Cross-domain few-shot learning results.}
\label{table:result}
\resizebox{\textwidth}{!}{
\begin{tabular}{lccc|ccc}
\toprule
\multirowcell{2}{Methods} & \multicolumn{3}{c|}{ChestX} & \multicolumn{3}{c}{ISIC} \\ 
\cline{2-7} 
 & 5-way 5-shot & 5-way 20-shot & 5-way 50-shot & 5-way 5-shot & 5-way 20-shot & 5-way 50-shot \\ 
\cline{1-7}
Fine-tuning~\cite{Guo2019A} & 25.99\%\(\pm\)0.42\% & 31.28\%\(\pm\)0.44\% & 36.78\(\pm\)0.49\% & 49.08\%\(\pm\)0.59\% & 58.98\%\(\pm\)0.56\% & 66.91\%\(\pm\)0.54\%\\
DFMN + MCT~\cite{kye2020transductive} & 24.69\%\(\pm\)0.63\% & 27.93\%\(\pm\)0.49\% & 31.60\(\pm\)0.46\% & 45.75\%\(\pm\)0.58\% & 51.83\%\(\pm\)0.52\% & 54.42\%\(\pm\)0.55\%\\
TMHFS & 26.20\%\(\pm\)0.44\% & 34.20\%\(\pm\)0.49\% & 39.55\(\pm\)0.53\% & 53.63\%\(\pm\)0.69\% & 65.46\%\(\pm\)0.48\% & 71.70\%\(\pm\)0.57\%\\
TMHFS+DA & 27.98\%\(\pm\)0.45\% & 37.11\%\(\pm\)0.49\% & 43.43\(\pm\)0.67\% & 53.84\%\(\pm\)0.68\% & 65.43\%\(\pm\)0.60\% & 71.29\%\(\pm\)0.78\%\\
 \bottomrule
\end{tabular}}
\resizebox{\textwidth}{!}{
\begin{tabular}{lccc|ccc}
\toprule
\multirowcell{2}{Methods} & \multicolumn{3}{c|}{EuroSAT} & \multicolumn{3}{c}{CropDiseases} \\ 
\cline{2-7} 
 & 5-way 5-shot & 5-way 20-shot & 5-way 50-shot & 5-way 5-shot & 5-way 20-shot & 5-way 50-shot \\ 
\cline{1-7}
Fine-tuning~\cite{Guo2019A} & 79.64\%\(\pm\)0.58\% & 88.19\%\(\pm\)0.54\% & 90.52\(\pm\)0.38\% & 86.97\%\(\pm\)0.63\% & 95.25\%\(\pm\)0.32\% & 97.37\%\(\pm\)0.22\%\\
DFMN + MCT~\cite{kye2020transductive} & 73.69\%\(\pm\)0.63\% & 82.08\%\(\pm\)0.59\% & 83.53\(\pm\)0.55\% & 89.84\%\(\pm\)0.58\% & 93.67\%\(\pm\)0.42\% & 94.58\%\(\pm\)0.34\%\\
TMHFS & 83.49\%\(\pm\)0.60\% & 91.05\%\(\pm\)0.42\% & 94.30\(\pm\)0.27\% & 93.58\%\(\pm\)0.44\% & 97.90\%\(\pm\)0.21\% & 98.96\%\(\pm\)0.14\%\\
TMHFS+DA & 85.34\%\(\pm\)0.55\% & 92.42\%\(\pm\)0.41\% & 95.63\(\pm\)0.32\% & 95.28\%\(\pm\)0.35\% & 98.51\%\(\pm\)0.17\% & 99.28\%\(\pm\)0.13\%\\
 \bottomrule
\end{tabular}}
\setlength{\tabcolsep}{12pt}	
\begin{tabular}{ccccc}
\toprule
\multirowcell{2}{Average over all tasks } & Fine-tuning~\cite{Guo2019A}& DFMN + MCT~\cite{kye2020transductive} & TMHFS & TMHFS+DA \\ 
\cline{2-5}
& 67.25\% \(\pm\) 0.48\% & 62.80\% \(\pm\) 0.53\% & 70.84\% \(\pm\) 0.44\% & 72.13\% \(\pm\) 0.47\% \\ 
 \bottomrule
\end{tabular}
\end{center}
\end{table*}
\begin{table}[h]
\begin{center}
\caption{Hyperparameters of the augmentation methods.}
\label{table:Augmentation}
\begin{tabular}{l|c}
\toprule
Augmentation & Hyperparameters \\ 
\hline 
 Scale (S) & pixel $\times$ pixel : 84 $\times$ 84 \\\hline 
 RandomResizedCrop (C) & pixel $\times$ pixel : 84 $\times$ 84 \\\hline 
 ImageJitter (J) & \tabincell{c}{Brightness:0.4\\ Contrast:0.4\\ Color:0.4} \\\hline 
 RandomHorizontalFlip (H) & Flip probability 50\%  \\\hline 
 RandomRotation (R) & 0 - 45 degrees	\\
 \bottomrule
\end{tabular}
\end{center}
\end{table}
\begin{table}[h]
\begin{center}
\caption{The augmentation choices for different target domains.}
\label{table:Augmentation_method}
\begin{tabular}{l|c}
\toprule
Dataset & Augmentation \\ 
\hline 
 \tabincell{c}{ISIC, EuroSAT,\\ CropDiseases} & \tabincell{c}{S + SJHR + SR + SJ +\\ SH + SJHR + SR + SJR +\\ SJH + SH} \\
 \hline
 ChestX & \tabincell{c}{S + SJH + C + CJ +\\ CH + CJH + C + CJ + \\CJH + CH} \\
 \bottomrule
\end{tabular}
\end{center}
\end{table}

We implemented the proposed TMHFS method 
in PyTorch 1.0, and trained the method on a Nvidia Tesla P40 GPU with 24Gb memory. 
We used ResNet-12 as the backbone network.
We trained the model with SGD optimizer with 50000 episodes at the training stage. 
We used C=15 during training and 
set the initial learning rate to 0.1 and cut it to 0.006 and 0.0012 at 25000 and 35000 episodes, respectively. 
At the fine-tuning stage, we set the epoch as 100 and learning rate as 0.01. 
For loss $L_S$ we set the batch size as 4 for training and fine-tuning. 
We set $(\lambda, \alpha)$ as $(0.2,0.4)$. 
For testing, we use the same 600 randomly sampled few-shot
episodes, and compute the average accuracy
and 95\% confidence interval, following the challenge evaluation procedure. 

We compared the proposed TMHFS method with a fine-tuning baseline~\cite{Guo2019A}  
and the base model DFMN+MCT~\cite{kye2020transductive} 
by using miniImageNet as the source domain and using ChestX, ISIC, EuroSAT and CropDiseases as
the target domain respectively. 
The results are presented in Table \ref{table:result}. 
We can see that fine-tuning is a strong baseline, while DFMN+MCT yields significant inferior results, 
and the proposed TMHFS consistently outperforms fine-tuning.

To further incorporate data augmentation techniques, 
we choose 5 types of augmentation methods, 
including  Scale (S), RandomResizedCrop (C), ImageJitter (J), RandomHorizontalFlip (H), and RandomRotation (R).
Their hyperparameters are shown in Table \ref{table:Augmentation}.
We use $n_A=10$ different combinations of augmentations to generate augmenting images.
The 10 compound augmentation methods we used for different target domains are 
shown in Table \ref{table:Augmentation_method}.
From Table \ref{table:result} we can see that with data augmentation
TMHFS+DA further improves the cross-domain few-shot learning performance
on ChestX, EuroSAT and CropDiseases. 

The average results across all datasets and shot levels 
have also been computed.
We can see that the average accuracy of TMHFS is 70.84\% (0.44), 
which is much higher than the results of Fine-tuning~\cite{Guo2019A} 67.25\% (0.48) 
and DFMN+MCT~\cite{kye2020transductive} 62.80\% (0.53).
TMHFS+DA further improves the result to 72.13\% (0.47).
These results verified the efficacy of the proposed approaches.

\section{Conclusion }
In this paper, we proposed a new Transductive Multi-Head Few-Shot (TMHFS) learning
method to address
cross-domain few-shot classification
and further improved it with data augmentations to yield TMHFS+DA. 
We conducted experiments on four target domains and the results show the proposed methods
greatly outperform the strong fine-tuning baseline and the standard transductive 
few-shot learning method DFMN+MCT.

{\small
\bibliographystyle{ieee}
\bibliography{egbib}

\begin{thebibliography}{1}\itemsep=-1pt

\bibitem{Guo2019A}
Y.~Guo, N.~C.~F. Codella, L.~Karlinsky, J.~R. Smith, T.~Rosing, and R.~Feris.
\newblock A new benchmark for evaluation of cross-domain few-shot learning.
\newblock 2019.

\bibitem{kye2020transductive}
S.~M. Kye, H.~B. Lee, H.~Kim, and S.~J. Hwang.
\newblock Transductive few-shot learning with meta-learned confidence.
\newblock {\em arXiv preprint arXiv:2002.12017}, 2020.

\end{thebibliography}
}
\end{document}